\documentclass[letterpaper, 10 pt, conference]{ieeeconf}  

\IEEEoverridecommandlockouts                              

\usepackage{graphics} 
\usepackage{epsfig} 
\usepackage{amsmath} 
\usepackage{amssymb}  
\usepackage{lipsum}
\usepackage{xcolor}
\usepackage{bm}
\usepackage{booktabs}

\usepackage[style=ieee, citestyle=numeric-comp, backend=biber]{biblatex}

\DeclareSourcemap{
  \maps{
    \map{
      \pertype{article}
      \step[fieldset=language, null]
      \step[fieldset=url, null]
      \step[fieldset=doi, null]
      \step[fieldset=issn, null]
      \step[fieldset=isbn, null]
      \step[fieldset=note, null]
      \step[fieldset=editor, null]
      \step[fieldset=urldate, null]
      \step[fieldset=file, null]
    }
  }
}
\DeclareSourcemap{
  \maps{
    \map{
      \pertype{inproceedings}
      \step[fieldset=language, null]
      \step[fieldset=url, null]
      \step[fieldset=doi, null]
      \step[fieldset=issn, null]
      \step[fieldset=isbn, null]
      \step[fieldset=note, null]
      \step[fieldset=editor, null]
      \step[fieldset=urldate, null]
      \step[fieldset=file, null]
    }
  }
}
\DeclareSourcemap{
  \maps{
    \map{
      \pertype{incollection}
      \step[fieldset=language, null]
      \step[fieldset=url, null]
      \step[fieldset=doi, null]
      \step[fieldset=issn, null]
      \step[fieldset=isbn, null]
      \step[fieldset=note, null]
      \step[fieldset=editor, null]
      \step[fieldset=urldate, null]
      \step[fieldset=file, null]
    }
  }
}

\title{\LARGE \bf
Bio-inspired Thrust-Assisted Trotting on a legged-Aerial Robot via Real-Time QP-MPC
}

\author{Shreyansh Pitroda$^{1}$, Eric Sihite$^{2}$, Alireza Ramezani$^{1*}$, Morteza Gharib$^{2}$
\thanks{This work was supported by the U.S. National Science Foundation (NSF) through the NSF CAREER Award No.\ 2340278.}
\thanks{$^{1}$The authors are with SiliconeSynapse Labs, 
the Department of Electrical Engineering, Northeastern University, USA. 
{\tt pitroda.s@northeastern.edu}}%
\thanks{$^{2}$The authors are with the Department of Aerospace Engineering, 
California Institute of Technology, USA. 
{\tt esihite, mgharib@caltech.edu}}%
\thanks{$^{*}$ The corresponding author. Email: {\tt \small a.ramezani@northeastern.edu}}%
}

\begin{document}

\maketitle
\thispagestyle{empty}
\pagestyle{empty}

\begin{abstract}
This paper presents the hardware implementation of a quadratic program-based controller on Harpy: thruster-assisted bipedal robot to perform thruster-assisted trotting. This is inspired by wing-assisted locomotion observed in birds such as the chukar partridge, which employ thrust vectoring to manipulate the ground reaction force. The proposed controller formulates the thruster-assisted spring-loaded inverted pendulum dynamics as a condensed QP-based model predictive control problem, enabling simultaneous regulation of ground contact force and active thrust vectoring at 500 Hz. A decoupled control architecture is used, where the robot's sagittal dynamics are controlled using QP-MPC while the frontal dynamics is regulated using a cascade orientation stabilization controller. Two experiments were conducted with different forward reference velocities of 0 m/s and 0.1 m/s. While the instantaneous velocity oscillates, the mean velocities can track their respective reference. Additionally, joint tracking error remains within $\pm 1^{\circ}$ while tracking the desired leg length given by the leg trajectory controller and desired swing trajectory given by Raibert's heuristic. The QP-MPC problem is consistently solved with 0.18ms, which is well within the 2ms computation budget while satisfying the GRF friction cone constraint.

\end{abstract}


\section{Introduction}

Birds are capable of using their wings to fly and also use their feet to walk on the ground, perch, and grabbing objects. Additionally, birds such as chukar’s wing-assisted incline running \cite{bundle_mechanics_2003} and widowbirds’ controlled flapping descent \cite{jackson_precocial_2009} can use their wings while walking over a steep slope or during a rapid descent to stabilize their upper body orientation and velocity. We aim to take inspiration from and mimic the biological counterparts by combining both ground and aerial locomotion modes to achieve a versatile legged-aerial multi-modal robot. 

There are several studies focused on multi-modal locomotion, specifically ground-aerial robots showing their versatility in traversing various terrains. Some of these studies focus more on transitioning between locomotion modes instead of using all available actuators at the same time. 
Some robot design requires transformation to switch mode of locomotion, such as \cite{sihite_multi-modal_2023,salagame_nmpc-based_2025}, which transitions between terrestrial and aerial locomotion by morphing its appendages,  \cite{wang_dynamic_2025} is a quadrupedal robot with propellers attached to its knee-axis and can perform flight after transforming into a quadrotor configuration.
On the other hand, some robot design with a fixed configuration such that they can perform ground and aerial locomotion without transforming, such as a jet powered bipedal robot capable performing large vertical jumps \cite{li_jet-hr2_2022}, pogo-stick mechanism with quadrotor dynamics \cite{zhu_pogodrone_2022}, a bipedal robot \cite{kim_bipedal_2021} and a quadrupedal robot \cite{krishnamurthy2025thruster} with quadrotor style propulsion attached to its torso.
\begin{figure}[t]
\vspace{0.07in}
    \centering
    \includegraphics[width=1\linewidth]{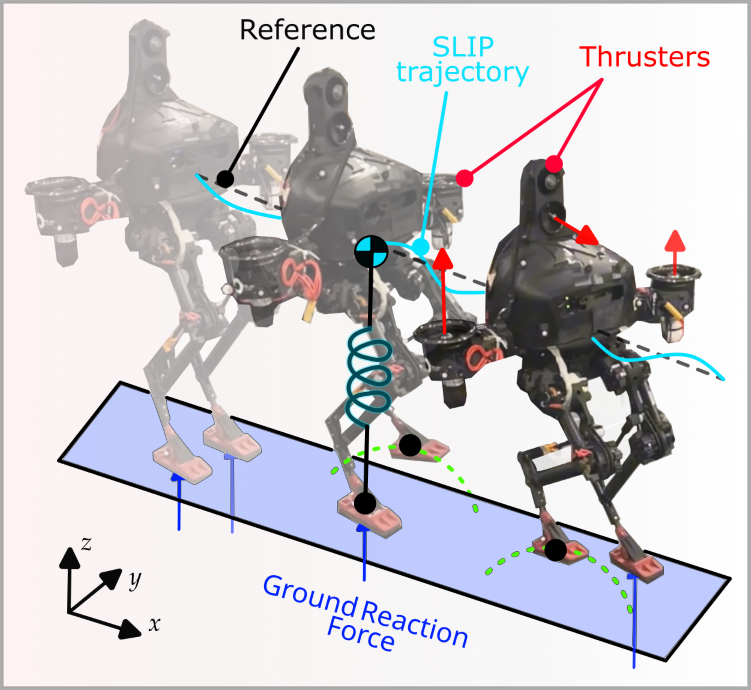}
    \caption{Illustrates biological inspiration for thruster-assisted bipedal robot. Birds employ wing-assisted locomotion to manipulate GRF~\cite{bundle_mechanics_2003}. Harpy replicates this strategy by replacing biological wings with electric ducted fans, enabling active thrust vectoring.}
    \label{fig:-cover}
\vspace{-0.07in}
\end{figure}
%
A multi-modal robot can utilize both ground and aerial actuators at the same time if the design allows it, making it possible to augment a mode of locomotion or perform a gait that can't be done by using either ground or aerial actuator by themselves. This work explores the capability of our legged-aerial Harpy platform, shown in Fig.~\ref{fig:-cover}, in performing a trotting gait using both the legs and thrusters at the same time. Bipedal robot locomotion requires a regulation of ground reaction force (GRF) so that the foot does not slip. The addition of thruster forces during walking will impact the resulting GRF and must be regulated together with the walking gait. Simulation studies have laid a theoretical groundwork for using both legs and thrusters at the same time to augment the legged locomotion to perform difficult maneuvers, such as narrow path walking \cite{wang2025quadratic,krishnamurthy_narrow-path_2024} and steep incline walking \cite{krishnamurthy_enabling_2024,krishnamurthy_optimization_2024,krishnamurthy2025thruster,venkatesh_krishnamurthy_wair_2026,krishnamurthy_contact-implicit_2026}. Building on this foundation, the Harpy platform has also been validated in simulation across various locomotion tasks, such as jumping~\cite{dangol_control_2021, sihite_posture_2024}, walking~\cite{pitroda2024capture,pitroda_enhanced_2024,pitroda_conjugate_2025,wang_thruster-enhanced_2025}, and inclined slope walking~\cite{pitroda_quadratic_2024}. However, unlike in simulations, the practical usage of the algorithm in a real-time controller requires an approach that can be solved quickly and efficiently.

A common solution for GRF regulation for legged robot uses an optimization method, such as quadratic programming (QP) or nonlinear model predictive control (MPC), by modeling the robot's dynamics and implementing the foot no-slip condition (the friction cone constraint) as constraints in the optimization. The optimization problem needs to be solved quickly to be feasible for a real-time implementation, where a reduced order model (ROM) is often used to reduce the model complexity and computational time. 

The reduced-order modeling tool, such as the Linear inverted pendulum (LIP) and the spring-loaded inverted pendulum (SLIP) model, captures the essential dynamics of walking or running~\cite{blickhan_spring-mass_1989,geyer_compliant_2006}. The SLIP model, first formalized by~\cite{blickhan_spring-mass_1989} and validated through extensive biomechanical studies~\cite{full_templates_1999,geyer_compliant_2006}, has emerged as the canonical template for understanding and controlling such dynamic gaits. The SLIP model's elegance lies in its simplicity: a point mass connected to a massless spring leg model, where the ground contact forces is applied alternatively between stance and flight phases. The reduced-order representation has proven remarkably effective in describing the locomotion of diverse running animals~\cite{holmes_dynamics_2006} and has guided the development of numerous robotic systems, from Raibert's pioneering hopper~\cite{raibert_experiments_1984} to modern dynamic bipeds~\cite{poulakakis_spring_2009,hubicki_walking_2018,gong_feedback_2019,dadashzadeh_slip-based_2020,wang_fast_2022,sihite_locomotion_2026,wang_legged_2023,ramezani2021generative,salagame_letter_2022}.

In this paper, we formulated an MPC optimization problem into a QP, and implemented it in our Harpy robot to satisfy ground constraints to perform a thruster-assisted trotting gait. The QP is formulated using a Thruster-SLIP (TSLIP) model, where thrusters are incorporated in the SLIP reduced order model, which then is implemented on hardware and solved in real-time using the qpSWIFT \cite{pandala_qpswift_2019} algorithm at 500 Hz control loop frequency. This controller is then tested on hardware during the thruster-assisted trotting gait, where the experimental results showed the controller's feasibility and performance. The contributions of this paper is listed as follows: 1) the derivation of the dynamic Thruster-SLIP model of a legged-aerial robot, 2) implementation of the QP-SLIP model on hardware, and 3) experimental validation of the controller performance showing the satisfaction of the GRF constraints in real-time. This paper is organized as follows: the robot design overview, followed by dynamic modeling, controller design and formulation, leg trajectory controller. Then, we present the experimental results, discussions, concluding remarks, and future work.

\section{Robot Overview}

Harpy~\cite{pitroda_dynamic_2023,dangol_performance_2020,dangol_feedback_2020} is designed with three joints for each leg: hip frontal joint, hip sagittal joint, and knee joint. The lower leg forms a parallelogram linkage, where a shock absorber spring is attached to the rear lower linkage to reduce the impulsive force on leg touchdown. There are four electric ducted fans (EDFs) attached to the torso: a high-powered EDF at each side of the robot, and two smaller upper EDFs with front/rear-facing orientation. Each larger and smaller EDF can generate a maximum thrust of 5kgf and 0.9kgf, respectively. The total weight of the robot is 7.1kg. Each joint is made with 3D printed material with embedded components. Joints are further connected using oval shape 2mm thickness carbon-fiber tube. The robot is equipped with STM32 F429ZI, which acts as a communication hub between ICM-20948 (IMU), OptiTrack data received over Wi-Fi using ESP32, and the xPC target machine.
\begin{figure}[t]
\vspace{0.07in}
    \centering
    \includegraphics[width=1\linewidth]{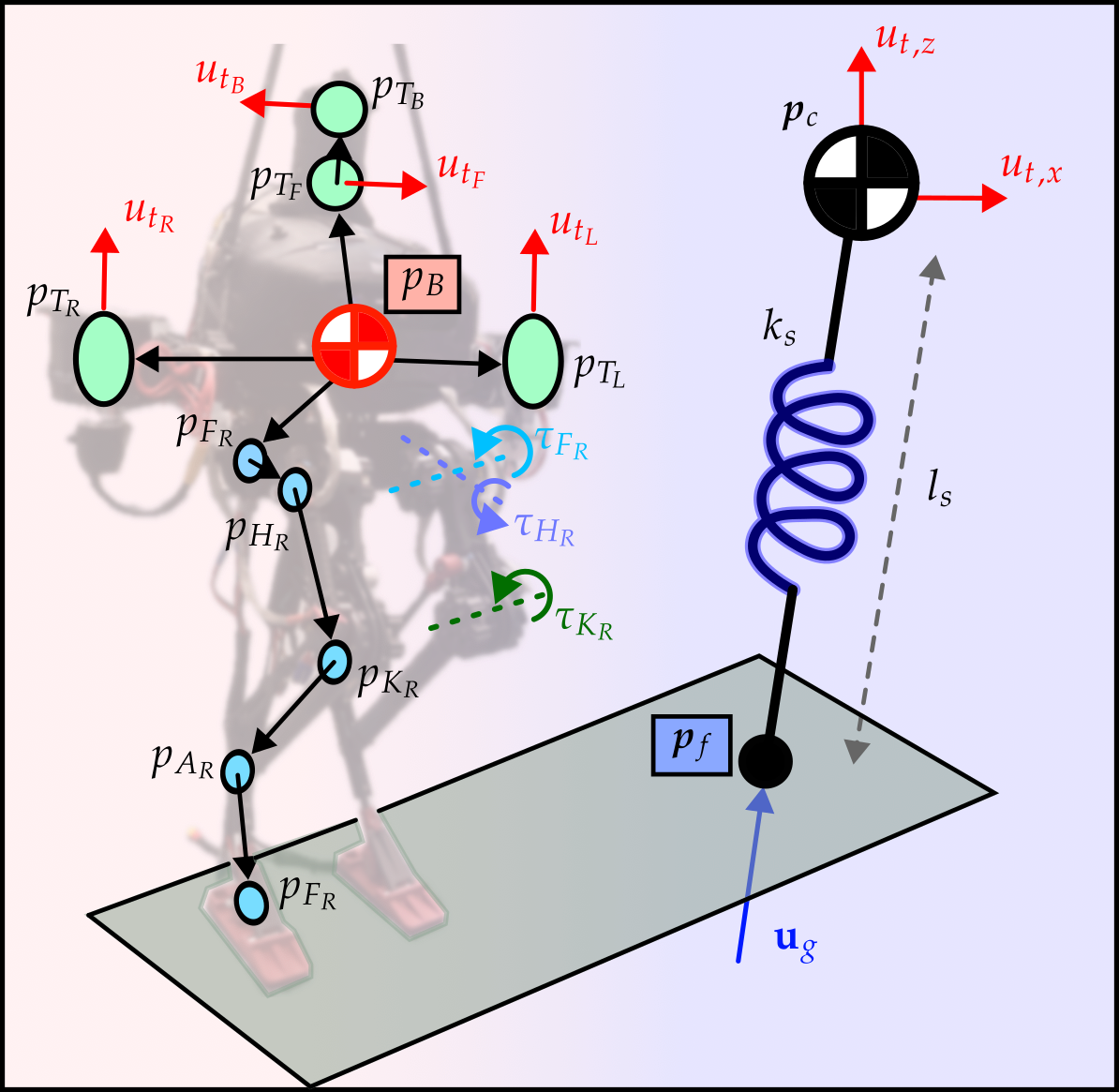}
    \caption{(Left) Full-order kinematic model of our robot highlighting joint, and thruster position. (Right) Planar 2D Spring-loaded Inverted Pendulum reduced-order model used for QP-MPC formulation.}
    \label{fig:-kinemaitics-SLIP-mdl}
\vspace{-0.07in}
\end{figure}
\section{Dynamic Modeling}

Consider the thruster-assisted bipedal robot shown in Fig.~\ref{fig:-kinemaitics-SLIP-mdl}. The high fidelity model of the system can be developed using Lagrangian mechanics on the unpinned model. The Lagrangian is defined as $\mathcal{L} = \mathcal{K} - \mathcal{V}$, where the kinetic energy $(\mathcal{K})$ and potential energy $(\mathcal{V})$ are given by
\begin{align*}
    \mathcal{K} &= \frac{1}{2}\sum_{i\in\mathcal{F}} \left( m_i\,\dot{\bm p}_i^\top \dot{\bm p}_i + \bm \omega_i^\top I_i\,\bm \omega_i \right),\\
    \mathcal{V} &= \sum_{i\in\mathcal{F}} m_i\, g\, p_{i,z},
\end{align*}
where $\mathcal{F} = \{B, F_L,F_R,H_L, H_R, K_L, K_R, A_L, A_R\}$ denotes the set of rigid bodies and links comprising of torso~$(B)$, frontal joints~$(F_L, F_R)$, hips~$(H_L, H_R)$, knees~$(K_L, K_R)$, and ankles~$(A_L, A_R)$. Here, $(L)$ and $(R)$ denote the left and right leg, respectively. For each body $i \in \mathcal{F}$, $m_i$ denotes the mass, $I_i \in \mathbb{R}^{3\times3}$ is the rotational inertia tensor expressed in the body frame, $\bm{p}_i \in \mathbb{R}^{3 \times 1}$ is the center-of-mass position for each link and body, $\bm \omega_i \in \mathbb{R}^{3 \times 1}$ is the angular velocity, and $g$ is the gravitational acceleration constant.

The dynamics can be derived by applying Hamilton's principle to the Lagrangian as:
\begin{equation}
M(\bm q)\ddot{\bm q} + \bm h(\bm q, \dot{\bm q}) = B_m \, \bm u_m + B_t(\bm q) \, \bm u_t + J_c^\top(\bm q)\,\bm \lambda_c,
\label{eq:eom}
\end{equation}
where the generalized coordinates are defined as:
\begin{equation}
\bm q = \begin{bmatrix} \bm p_B^\top & \bm \Theta_B^\top & \bm q_j^\top \end{bmatrix}^\top \in \mathbb{R}^{n_q},
\label{eq:gen_coords}
\end{equation}
with $n_q = 12$, $\bm p_B \in \mathbb{R}^3$ denoting the torso position, $\bm \Theta_B \in \mathbb{R}^3$ representing the body orientation parameterized using Euler angles in the XYZ (roll-pitch-yaw) convention, and $\bm q_j \in \mathbb{R}^{6}$ containing the six joint angles. Further, the terms in Eq.~\eqref{eq:eom} are defined where $M(\bm q) \in \mathbb{R}^{n_q \times n_q}$ is the symmetric positive-definite mass-inertia matrix, $\bm h(\bm q, \dot{\bm q}) = C(\bm q, \dot{\bm q})\,\dot{\bm q} + G(\bm q)$ is the vector combining Coriolis/centrifugal forces and gravitational forces, $B_m \in \mathbb{R}^{n_q \times 6}$ is the joint torque selection matrix, $\bm u_m \in \mathbb{R}^6$ is the vector of joint torques, $\bm u_t \in \mathbb{R}^{n_t}$ is the vector of thruster forces with $n_t = 4$ and $\bm \lambda_c \in \mathbb{R}^{n_c}$ is the ground reaction forces (GRF) at the contact point with $n_c \le 6$ depending on contact configuration.

The contact Jacobian $J_c(\bm q) \in \mathbb{R}^{n_c \times n_q}$ maps generalized velocities to contact point velocities and is constructed as:
\begin{equation}
J_c(\bm q) = \begin{bmatrix}
    J_{F_L}(\bm q) \\
    J_{F_R}(\bm q)
\end{bmatrix}, \quad \text{where} \quad J_{F_i}(\bm q) = \frac{\partial \bm p_{F_i}}{\partial \bm q},
\label{eq:contact_jac}
\end{equation}
with $\bm p_{F_L}, \bm p_{F_R} \in \mathbb{R}^3$ denoting the positions of the left and right foot contact points, respectively. The thruster Jacobian $B_t(\bm q) \in \mathbb{R}^{n_q \times n_t}$ maps thruster forces to generalized forces:
\begin{equation}
B_t(\bm q) = \begin{bmatrix}
    J_{T_L}^\top(\bm q) & J_{T_R}^\top(\bm q) & J_{T_F}^\top(\bm q) & J_{T_B}^\top(\bm q)
\end{bmatrix}^\top,
\label{eq:thruster_jac}
\end{equation}
where $J_{T_i}(\bm q) = \partial \bm p_{T_i} / \partial \bm q$ for each thruster mounting point $T_i \in \{T_L, T_R, T_F, T_B\}$, representing the left, right, front-facing, and back-facing thrusters. $B_t(\bm q)$ is configuration-dependent since it is applied in the body-fixed frame and its effective contribution to the generalized coordinate varies with the body's orientation.
\begin{table}[t]
\centering
\caption{Mapping PWM to thruster forces}
\label{tab:thrust-table}
\begin{tabular}{ccc}
\toprule
\textbf{PWM} & \textbf{Side thruster} & \textbf{Upper thruster} \\
\textbf{(\%)} & \textbf{(kgf)} & \textbf{(kgf)} \\
\midrule
10 & 0.092 & 0.0875 \\
20 & 0.37 & 0.175 \\
30 & 0.809 & 0.2625 \\
40 & 1.346 & 0.35 \\
50 & 1.97 & 0.4375 \\
60 & 2.674 & 0.5250 \\
70 & 3.33 & 0.6125 \\
80 & 3.91 & 0.7 \\
90 & 4.49 & 0.78 \\
100 & 5.25 & 0.87 \\
\bottomrule
\end{tabular}
\end{table}
\section{Decoupled controller formulation}
\begin{figure}[t]
\vspace{0.07in}
    \centering
    \includegraphics[width=1\linewidth]{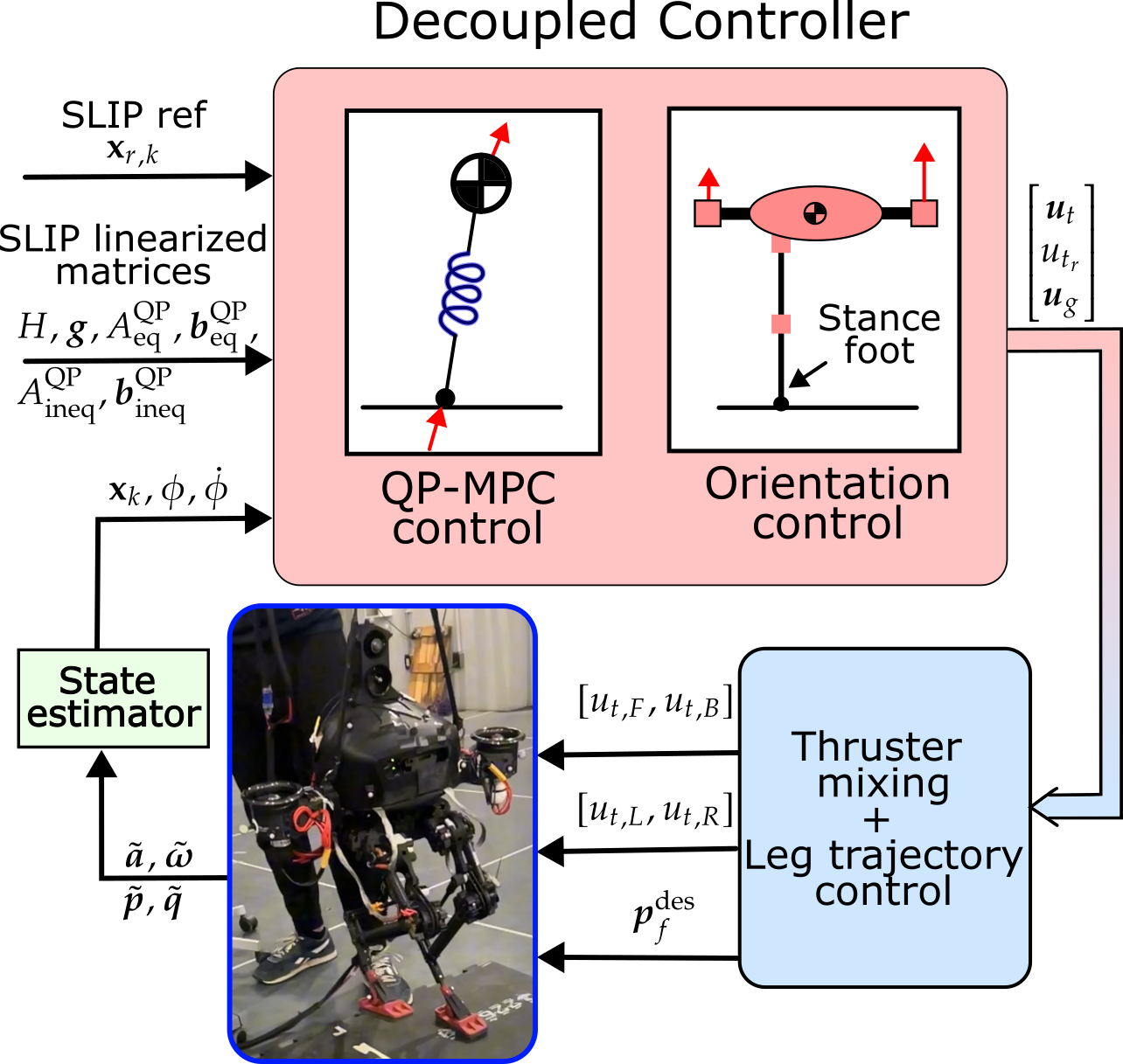}
    \caption{Control architecture of the proposed QP-TSLIP controller.}
    \label{fig:-con-arch}
\vspace{-0.07in}
\end{figure}

The overall control architecture is illustrated in Fig.~\ref{fig:-con-arch}. The proposed framework adopts a decoupled structure separating the sagittal locomotion problem, handled by the TSLIP-based QP-MPC, from the frontal-plane stabilization problem, handled by the orientation controller. 
\subsection{ROM dynamics}
The ROM considers the mass $(m_b)$ concentrated at single point defined by $\bm{p}_c = [p_{c,x},p_{c,z}]^{\top}$ and ignores the orientation of the robot. Two different forces are acting on the point mass, 
\begin{equation}
\mathbf{u}_t = [u_{t,x},u_{t,z}], \qquad \mathbf{u}_g = [u_{g,x},u_{g,z}],
\end{equation}
where, $\mathbf{u}_t$ is the thrust force acting at the COM position, and $\mathbf{u}_g$ is the GRF acting at the point of contact $\bm{p}_f = [p_{f,x},p_{f,z}]^{\top}$. Under these forces, the ROM dynamics are described as follows
\begin{equation}
    \dot{\mathbf{x}} = A~\mathbf{x} + B~\mathbf{u} - [0,0,0,g]^{\top},
    \label{eq:ROM-dyn}
\end{equation}
where, state vector is defined as $\mathbf{x} = [\bm{p}_c^{\top}, \dot{\bm{p}}_c^{\top}]^{\top} \in \mathbb{R}^4$, input vector is defined as $\mathbf{u} = [\bm u^{\top}_{g}, \bm u^{\top}_{t}]^{\top} \in \mathbb{R}^4$ and the system matrices follows directly from Newton's second law applied to the point mass:
\begin{equation}
  A = \begin{bmatrix} 0 & 0 & 1 & 0 \\ 0 & 0 & 0 & 1 \\
                      0 & 0 & 0 & 0 \\ 0 & 0 & 0 & 0 \end{bmatrix},
  \quad
  B = \begin{bmatrix} 0 & 0 & 0 & 0 \\ 0 & 0 & 0 & 0 \\
                      \tfrac{1}{m_b} & 0 & \tfrac{1}{m_b} & 0 \\
                      0 & \tfrac{1}{m_b} & 0 & \tfrac{1}{m_b} \end{bmatrix}.
\end{equation}
Applying the forward Euler method with step size $h$ gives
\begin{equation}
  \mathbf{x}_{i+1}
    = A_d\,\mathbf{x}_i
      + B_d\,\mathbf{u}_i
      + \mathbf{d}_d,
  \label{eq:rom_dt}
\end{equation}
where $i$ is the current time step, the terms of the discrete dynamics are defined as $A_d = (\mathbb{I} + Ah)$, $B_d = Bh$, and $\mathbf{d}_d = [0,0,0,-hg]^{\top}$.
\subsection{SLIP dynamics and constraint}
\label{sec:SLIP}
In the stance phase, the GRF is generated by a spring-loaded virtual leg connecting the COM to the foot contact point, which is held fixed by friction. The leg vector and length are
\begin{equation}
  \mathbf{r} = \bm{p}_c - \bm{p}_f, \qquad
  l_s = \|\mathbf{r}\|_2.
\end{equation}
With rest length $l_0$ and spring stiffness $k_s$, the spring force magnitude obeys a unilateral Hooke's law:
\begin{equation}
  F_g = -k_s\,(l_s - l_0)_{+},
\end{equation}
where $(\cdot)_{+} = \max(\cdot,\, 0)$ enforces that the spring can only push, not pull.  The corresponding GRF vector directed along the spring axis is
\begin{equation}
  \mathbf{u}_g = F_g\,\hat{\mathbf{e}}_\theta,
  \qquad \hat{\mathbf{e}}_\theta = \mathbf{r}/l_s.
  \label{eq:slip_grf}
\end{equation}

Since \eqref{eq:slip_grf} is nonlinear in the state, we linearize it about a nominal trajectory $(\bar{\mathbf{x}},\bar{\mathbf{u}})$ via a first-order Taylor expansion:
\begin{equation}
  \mathbf{u}_g \approx
    \mathbf{u}_g(\bar{\mathbf{x}},\bar{\mathbf{u}})
    + \left.\frac{\partial \mathbf{u}_g}{\partial \mathbf{x}}\right|_{\bar{\mathbf{x}},\bar{\mathbf{u}}}
      (\mathbf{x} - \bar{\mathbf{x}})
    + \left.\frac{\partial \mathbf{u}_g}{\partial \mathbf{u}}\right|_{\bar{\mathbf{x}},\bar{\mathbf{u}}}
      (\mathbf{u} - \bar{\mathbf{u}}).
  \label{eq:slip_linearized}
\end{equation}
This affine approximation can be compactly written as the equality constraint
\begin{equation}
  A_\text{eq}\,\mathbf{x} + B_\text{eq}\,\mathbf{u}
    + \mathbf{d}_\text{eq} = \mathbf{0}.
  \label{eq:eq_constraint}
\end{equation}
\subsection{QP-MPC formulation}
\label{sec:qp}
The SLIP equality constraint \eqref{eq:eq_constraint} is
embedded within a receding-horizon MPC framework that optimizes the states and the control inputs.  Given a reference trajectory $\{\mathbf{x}_{r,k}\}_{k=0}^{N}$ over prediction horizon $N$, positive-definite weight matrices $Q$ and $R$, and the initial condition $\mathbf{x}_\text{init}$, the MPC problem is
\begin{equation}
\begin{aligned}
  \min_{\{\mathbf{x}_k,\, \mathbf{u}_k\}} \quad
    & \sum_{k=0}^{N} \|\mathbf{x}_k - \mathbf{x}_{r,k}\|_Q^2
      + \sum_{k=0}^{N-1} \|\mathbf{u}_k\|_R^2 \\
  \text{s.t.} \quad
    & \mathbf{x}_{k+1} = A_d\,\mathbf{x}_k + B_d\,\mathbf{u}_k
        + \mathbf{d}_d, \\
    & \mathbf{x}_0 = \mathbf{x}_\text{init}, \\
    & A_\text{ineq}\,\mathbf{u}_k \le \mathbf{b}_\text{ineq}, \\
    & A_\text{eq}\,\mathbf{x}_k + B_\text{eq}\,\mathbf{u}_k
        + \mathbf{d}_\text{eq} = \mathbf{0}.
\end{aligned}
\label{eq:mpc}
\end{equation}
The input bounds enforce physical limits on the GRF and thruster forces.  The friction cone requires
\begin{align*}
  u_{g,z} \ge 0, \qquad |u_{g,x}| \le \mu\, u_{g,z},
\end{align*}
and the saturation limits for the thruster and GRF are
\begin{align*}
  & u_{t,x,\min} \le u_{t,x} \le u_{t,x,\max}, \qquad
  & 0 \le u_{t,z} \le u_{t,z,\max},\\
  & u_{g,x,\min} \le u_{g,x} \le u_{g,x,\max}, \qquad
  & 0 \le u_{g,z} \le u_{g,z,\max}.
\end{align*}
These are assembled into the block-diagonal inequality matrices $A_\text{ineq}$ and $\mathbf{b}_\text{ineq}$ over the full horizon.
To reduce the decision variables and improve solver speed, we eliminate the state from the decision variables through recursive forward simulation of~\eqref{eq:rom_dt}. Stacking states and inputs column-wise as 
\begin{equation}
 \mathbf{X}
    = \begin{bmatrix} \mathbf{x}_0^\top & \cdots & \mathbf{x}_N^\top \end{bmatrix}^\top,
  \quad
  \mathbf{U}
    = \begin{bmatrix} \mathbf{u}_0^\top & \cdots & \mathbf{u}_{N-1}^\top \end{bmatrix}^\top,
\end{equation}
The rollout over the horizon yields the affine map
\begin{equation}
    \mathbf{X} = \mathcal{M}\,\mathbf{x}_0 + \mathcal{N}\,\mathbf{U} + \mathcal{W},
    \label{eq:condense_dyn}
\end{equation}
where,
\begin{align}
  \mathcal{M}
    &= \begin{bmatrix} \mathbb{I} \\ A_d \\ \vdots \\ A_d^N \end{bmatrix},
  \quad
  \mathcal{N}
    = \begin{bmatrix}
        \mathbf{0}        & \cdots & \mathbf{0} \\
        B_d               & \cdots & \mathbf{0} \\
        \vdots            & \ddots & \vdots     \\
        A_d^{N-1}B_d      & \cdots & B_d
      \end{bmatrix},
  \\
  \mathcal{W}
    &= \begin{bmatrix}
        \mathbf{0} &
        \mathbf{d}^{\top}_d &
        \dots &
        (\ \sum_{i=0}^{N-1} A_d^i \mathbf{d}_d )\ ^{\top}
      \end{bmatrix}^{\top}.
\end{align}
Substituting~\eqref{eq:condense_dyn} into the SLIP equality constraints~\eqref{eq:eq_constraint} eliminates~$\mathbf{X}$, yielding a standard QP in $\mathbf{U} \in \mathbb{R}^{N_{n_u}}$: 
\begin{equation}
\begin{aligned}
  \min_{\mathbf{U}} \quad
    & \mathbf{U}^\top H\,\mathbf{U} + \mathbf{g}^\top\mathbf{U} \\
  \text{s.t.} \quad
    & A_\text{eq}^\text{QP}\,\mathbf{U} = \mathbf{b}_\text{eq}, \quad
      A_\text{ineq}^\text{QP}\,\mathbf{U}  \le \mathbf{b}_\text{ineq},
\end{aligned}
\label{eq:qp_condensed}
\end{equation}
with cost matrices
\begin{align*}
  H &= \mathcal{N}^\top \bar{Q}\,\mathcal{N} + \bar{R}, \\
  \mathbf{g} &= \mathcal{N}^\top \bar{Q}
               (\mathcal{M}\,\mathbf{x}_0 + \mathcal{W} - \mathbf{X}_r ),
\end{align*}
where $\bar{Q} = diag([Q,\cdots,Q]) \in \mathbb{R}^{(N+1)n_x \times (N+1)n_x}$, $\bar{R} = diag([R,\cdots,R]) \in \mathbb{R}^{N\,n_u \times N\,n_u}$ and $\mathbf{X}_r = [\mathbf{x}^{\top}_{r,0},\cdots,\mathbf{x}^{\top}_{r,N}]^{\top}$ is the stacked reference trajectory over the horizon.
\subsection{Roll Orientation Controller}
\label{sec:orient-cont}
While the sagittal dynamics are handled by the QP-MPC described above, roll stabilization during single-support is decoupled and delegated to the side thrusters. During the single-support phase, since robot doesn't have any ankle roll joint, the side thrusters provide the necessary actuation authority in the frontal plane.

A cascaded control architecture is employed. The outer loop is a PD controller that maps the COM lateral position error and velocity to a reference roll angle: 
\begin{equation}
    \phi_{\text{ref,k}} = -K_{p,y}\,e_y - K_{d,y}\,\dot{e}_y,
\end{equation}
where $e_y = y - y_{\text{in}}$ is the error in the lateral position from the initial position at the start of stance phase and $\dot{e}_y$ is the velocity error, and $K_{p,y},K_{d,y} > 0$ are tuned gains. The inner loop is a PID controller that tracks $\phi_{\text{ref}}$ by commanding side thruster force:
\begin{equation}
    \begin{aligned}
        u_{t_r,i} = - K_{p,\phi}\,(\phi - \phi_{\text{ref,k}}) &- K_{d,\phi}\,\dot{\phi} - K_{I,\phi}\int(\phi - \phi_{\text{ref,k}})\,\delta t \\&+ u_{t,\text{ff}},
    \end{aligned}
\end{equation}
where $i = [L,R]$ depends on leg that is currently on swing cycle, $\delta t = 0.002~\text{s}$ is the control loop rate, and $K_{p,\phi},K_{d,\phi}, K_{I,\phi} > 0$ are the gains. The feedforward term $u_{t,\text{ff}}$ applies a bias force to pre-compensate for the lateral moment that is induced by the swing leg movement. The roll angle $\phi$ is given by the onboard EKF, and $\dot{\phi}$ is given by the onboard complementary filter.
\section{Stance leg trajectory tracking controller}
\label{sec:leg-cont}
The stance leg controller translates the QP-MPC output into joint-space commands for the stance leg. At each control cycle, the optimal output $\mathbf{u}^{*}$ from QP are used to propagate the discrete dynamics~\eqref{eq:rom_dt} by one step, yielding a desired COM state:
\begin{align}
    \mathbf{x}^{\text{des}}_{k+1} = A_d\,\mathbf{x}_{k} + B_d\,\mathbf{u}^* + \mathbf{d}_d
\end{align}
The desired COM position $\bm{p}^{des}_c$ is extracted from $\mathbf{x}^{\text{des}}_{k+1}$. The corresponding desired foot position in the body frame is computed as follows 
\begin{align}
    \bm{p}^{des}_f = R_b(\bm{p}^{des}_c - \bm{p}_f),
\end{align}
where $\bm{p}_f$ is the current stance foot position in the inertial frame and $R_b \in SO(3)$ is the rotation matrix from the inertial to the body frame. The desired foot is then mapped to joint angles via inverse kinematics of the stance leg.
\section{Result}
\begin{figure}[t]
\vspace{0.07in}
    \centering
    \includegraphics[width=1\linewidth]{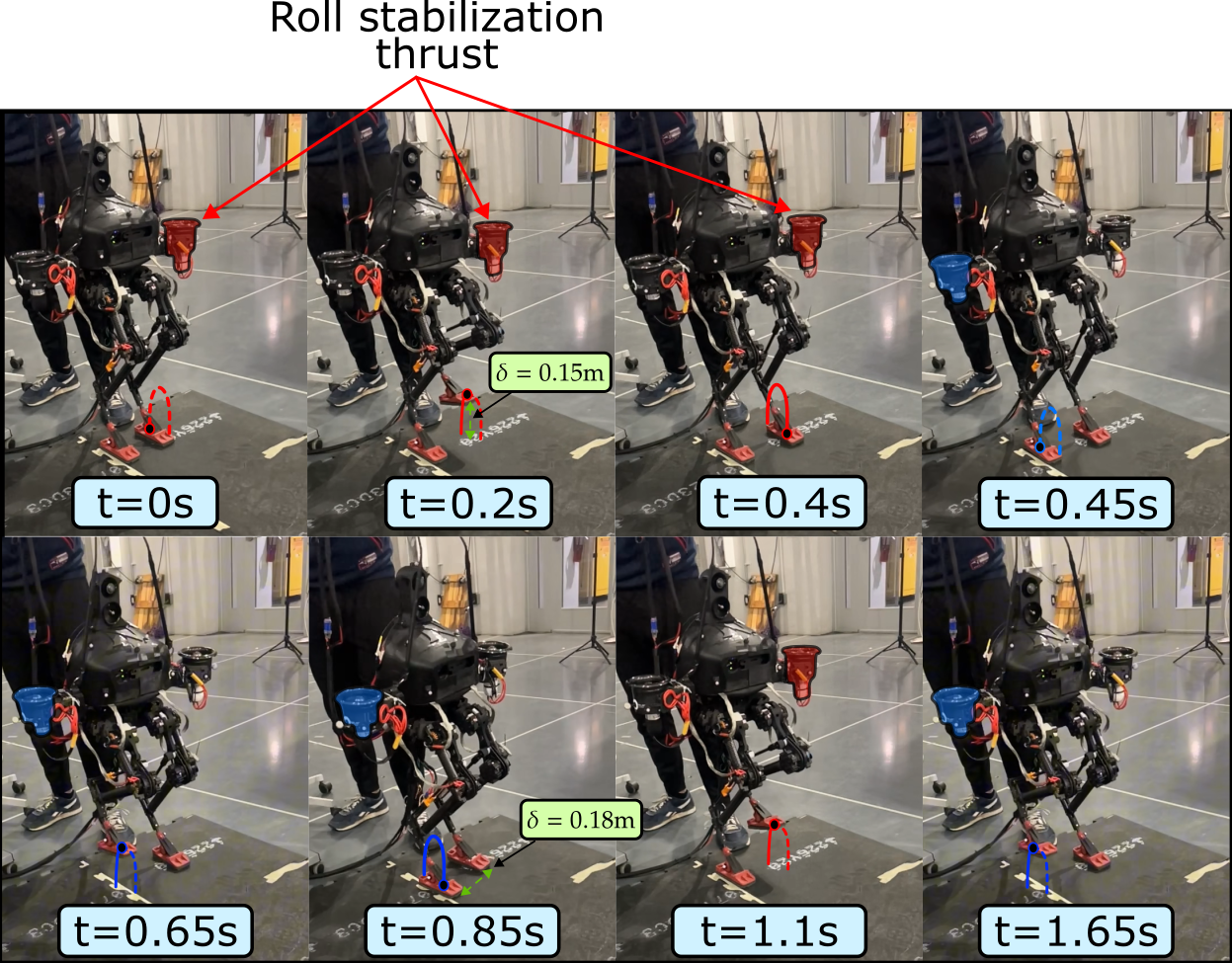}
    \caption{Sequential snapshots from hardware Experiment B. The robot performs thruster-assisted trotting locomotion using the proposed QP-TSLIP controller.}
    \label{fig:-hardware-trot}
\vspace{-0.07in}
\end{figure}
\begin{figure}[t]
\vspace{0.07in}
    \centering
    \includegraphics[width=1\linewidth]{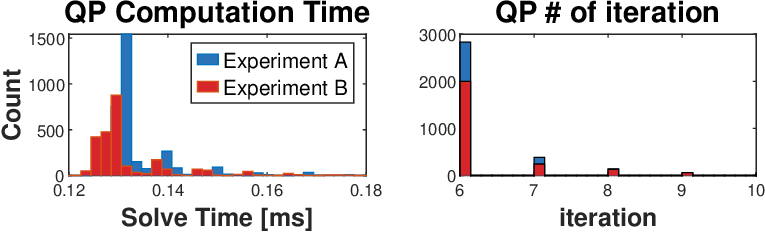}
    \caption{Illustrates QP solver performance on hardware. (Left) Histogram of QP total computation time for Experiment A (blue) and Experiment B (red). (Right) Distribution of solver iteration, displaying majority convergence within 6-7 steps.}
    \label{fig:-qp_solver_info}
\vspace{-0.07in}
\end{figure}
\begin{figure}[t]
\vspace{0.07in}
    \centering
    \includegraphics[width=1\linewidth]{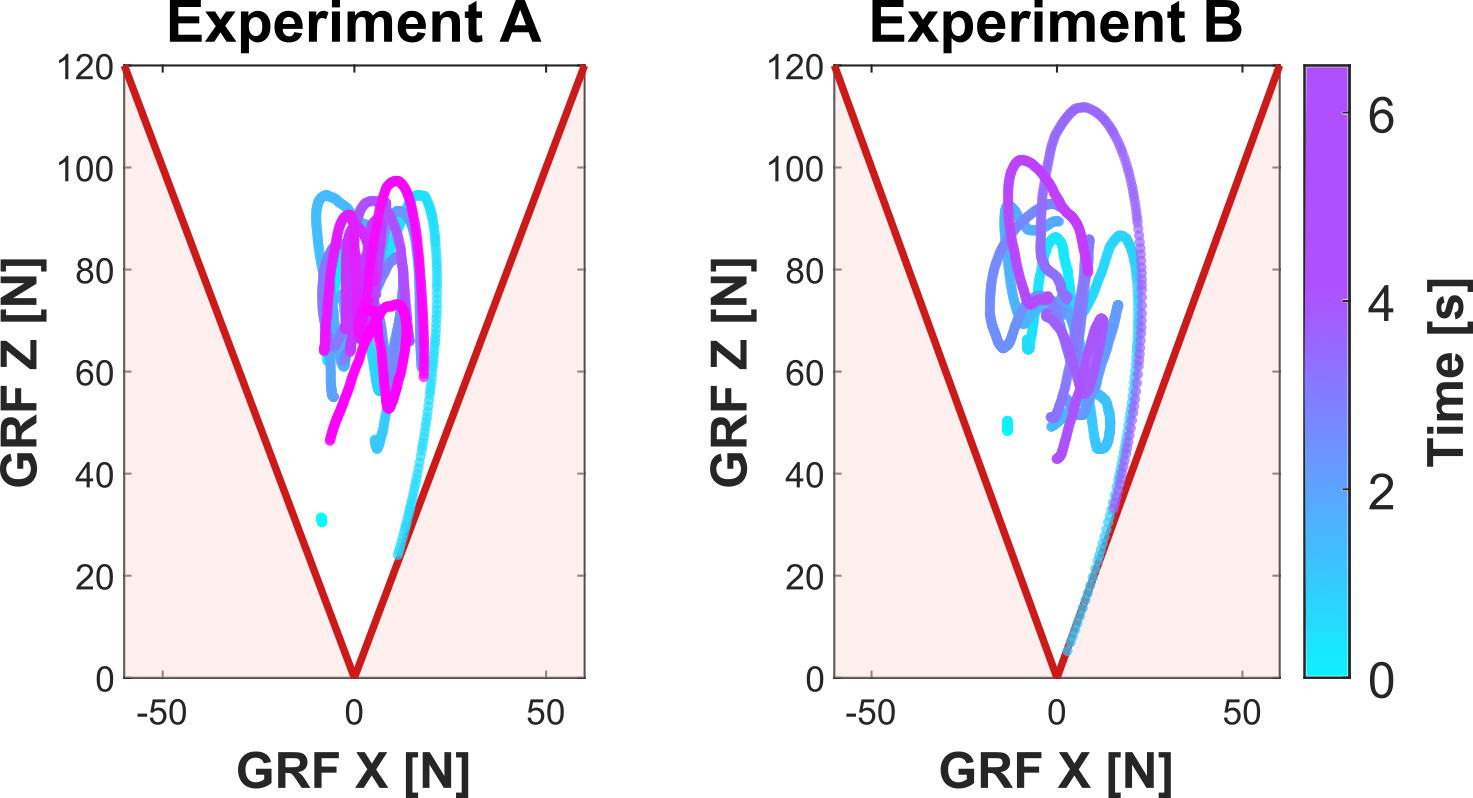}
    \caption{Illustrates GRF distribution within the friction cone for Experiment A (Left) and Experiment B (Right). Each point represents the horizontal and vertical force commanded by the QP solver, colored by time (cyan = start and magenta = end). The red line denotes the friction cone boundary with coefficient $\mu = 0.5$.}
    \label{fig:-grf}
\vspace{-0.07in}
\end{figure}
\begin{figure}[t]
\vspace{0.07in}
    \centering
    \includegraphics[width=1\linewidth]{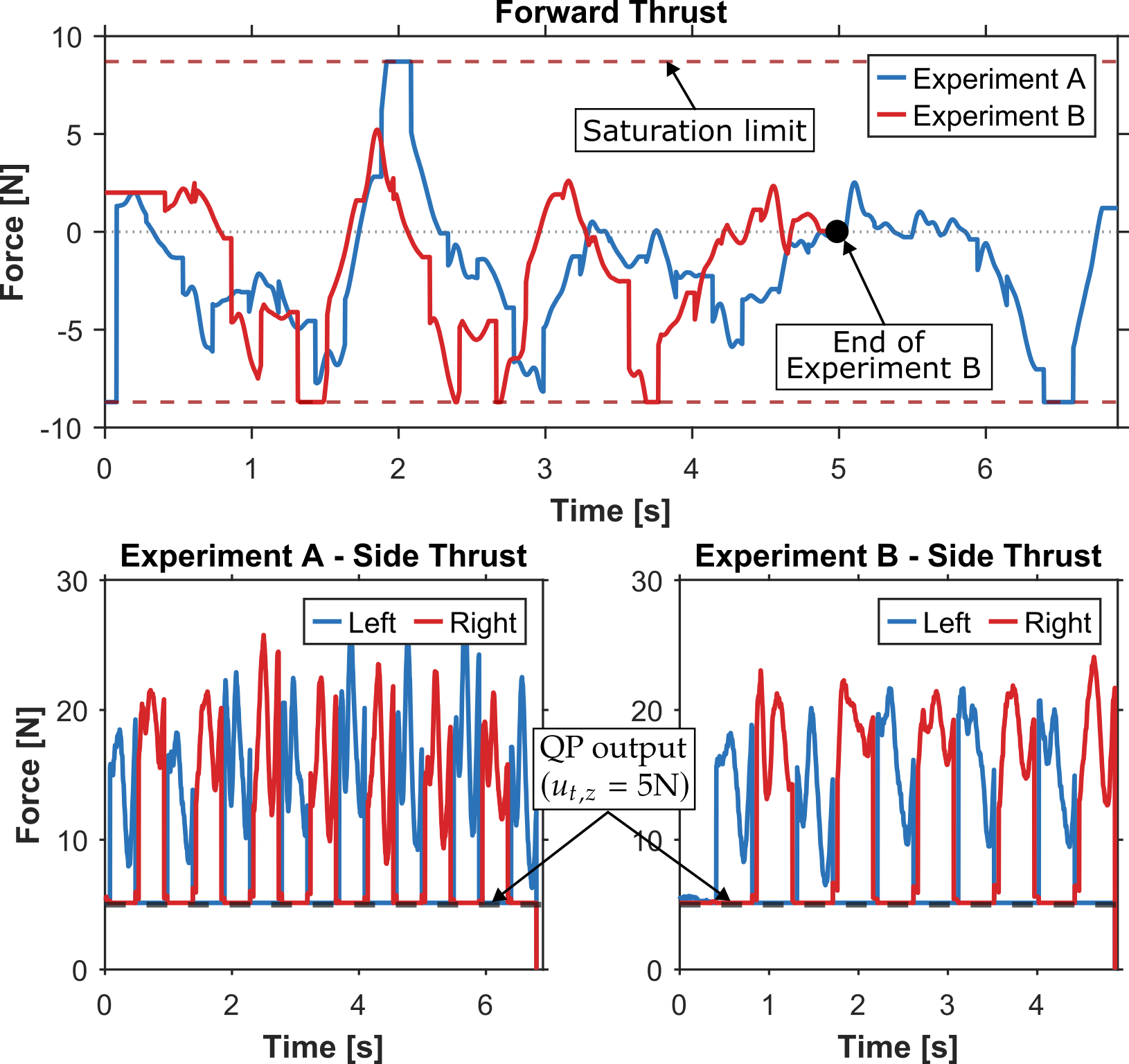}
    \caption{Thruster force commands during hardware experiments. (Top) QP-commanded forward thrust$~u_{t,x}$ for both the experiments. The thrust frequently remains near the saturation limit of $\pm 8.7~\text{N}$. (Bottom-left) Side thruster forces for Experiment A and (Bottom-right) for Experiment B, showing left and right thruster output combining PID orientation controller and QP controller. The dashed line at 5N denotes the baseline produced by the QP controller.}
    \label{fig:-thrust}
\vspace{-0.07in}
\end{figure}
\begin{figure}[t]
\vspace{0.07in}
    \centering
    \includegraphics[width=1\linewidth]{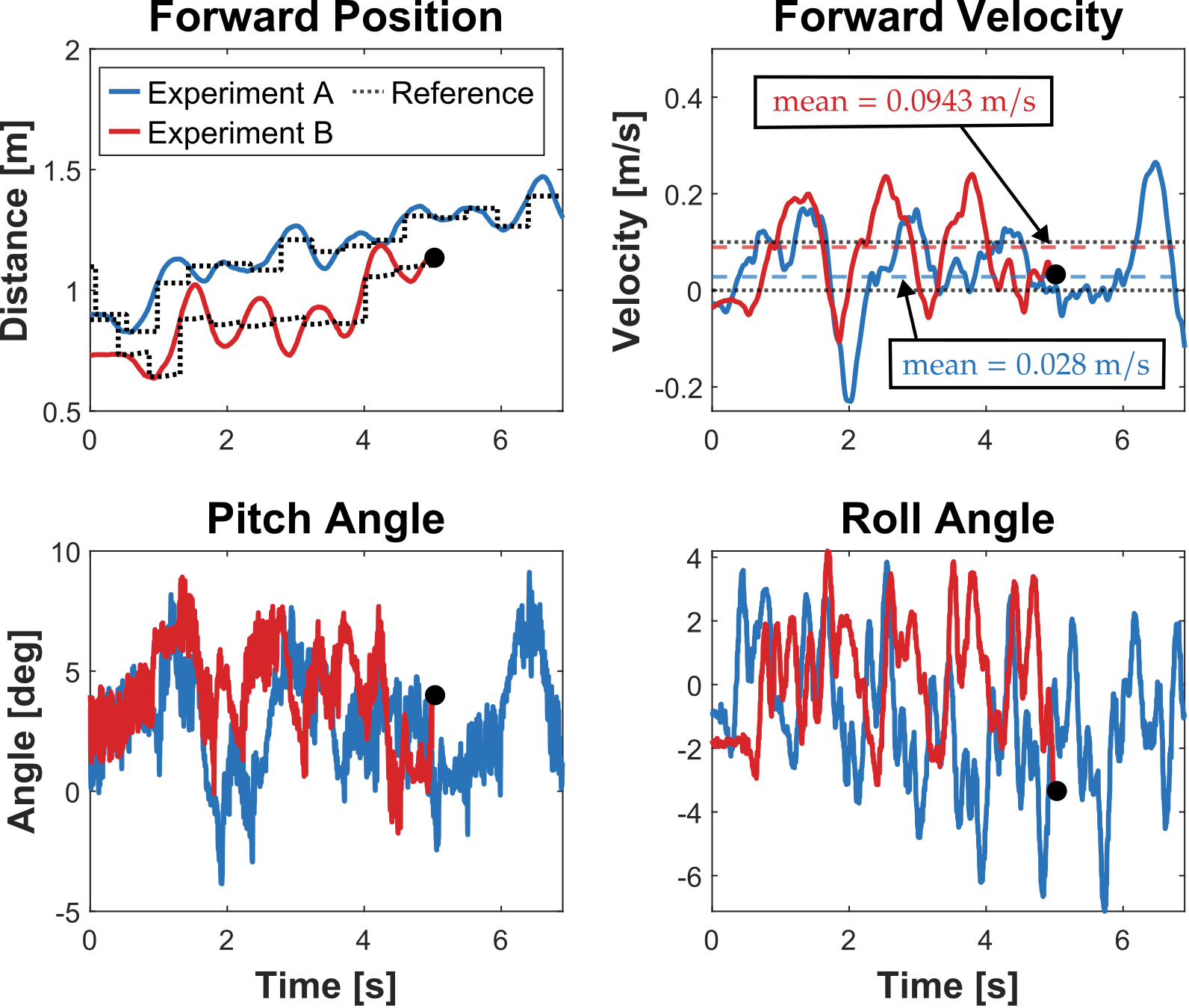}
    \caption{Body state trajectory during hardware experiments. (Top-left) COM position in x-direction for Experiment A (blue) and Experiment B (red) with dotted position reference line (black). (Top-right) body velocity in the x-direction, along with the mean values and velocity reference. (Bottom-left) Pitch angle remains bounded within $\pm 9^{\circ}$ throughout both experiments. (Bottom-right) Roll angle stays within $\pm 4^{\circ}$, validating the effectiveness of the cascade control strategy. The filled black circle $(\bullet)$ denotes the end of Experiment B.}
    \label{fig:-states}
\vspace{-0.07in}
\end{figure}
\begin{figure}[t]
\vspace{0.07in}
    \centering
    \includegraphics[width=1\linewidth]{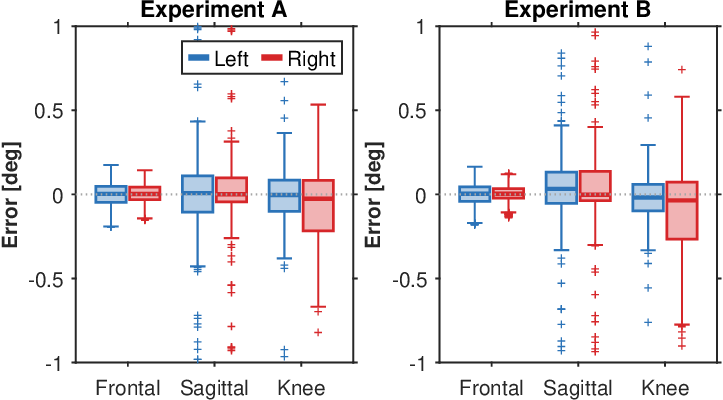}
    \caption{Box plot for joint trajectory tracking error for Experiment A and Experiment B. }
    \label{fig:-traj-track-err}
\vspace{-0.07in}
\end{figure}
%
%
\begin{figure}[t]
\vspace{0.07in}
    \centering
    \includegraphics[width=1\linewidth]{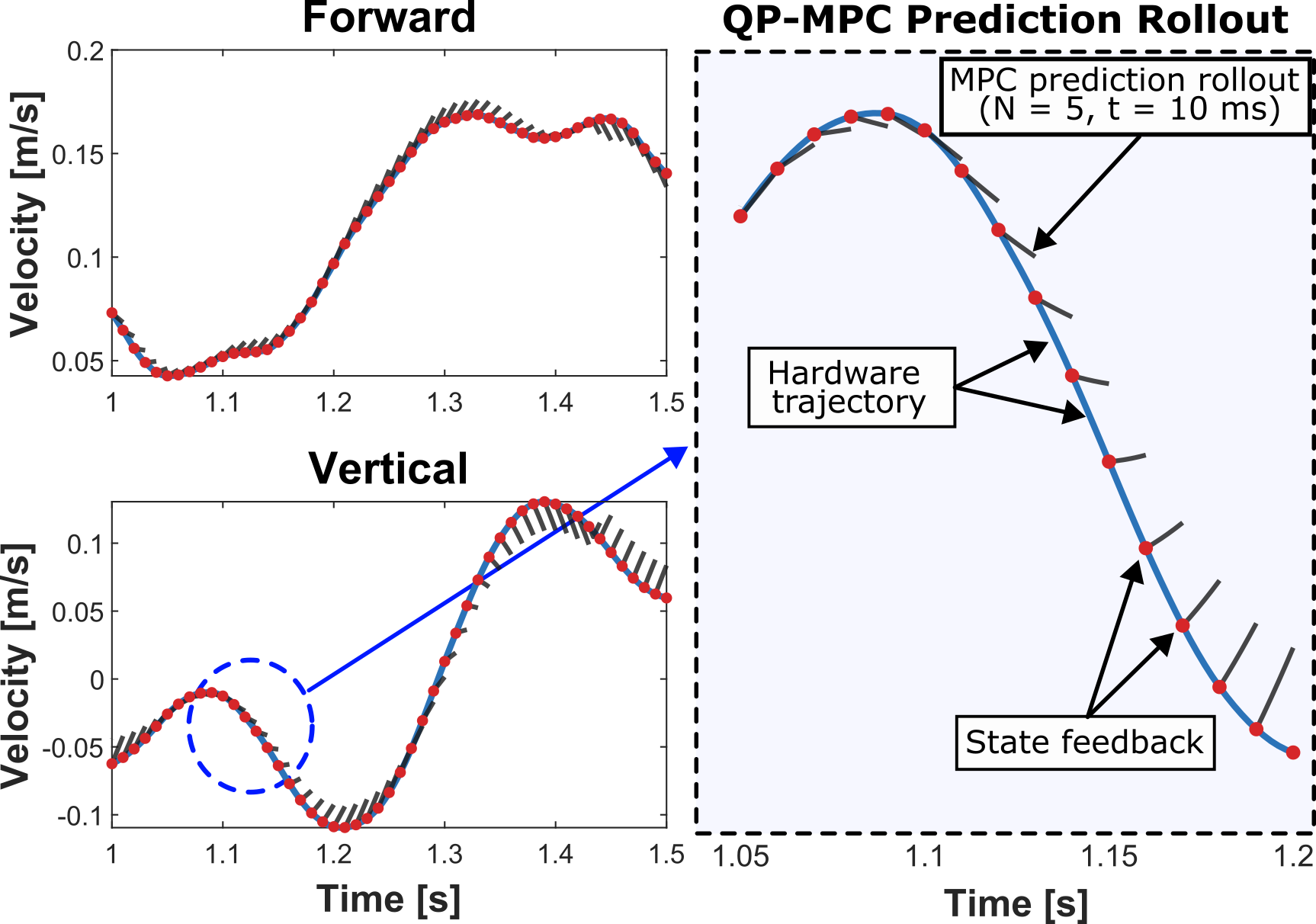}
    \caption{MPC prediction rollout validated on hardware for Experiment A. Forward (top-left), and vertical (Bottom-left) COM velocity trajectory comparing the QP-MPC model prediction against the hardware measurement (blue). The zoomed section illustrates the receding horizon strategy where at each state feedback point (red dot), the linearized SLIP model rollout (black) are calculated for 5 horizon.}
    \label{fig:-mpc-prediction}
\vspace{-0.07in}
\end{figure}
\subsection{Experimental Details}
Two hardware experiments were conducted: Experiment A with a forward velocity reference of 0 m/s over 15 steps and Experiment B with a velocity reference of 0.1 m/s over 10 steps. All experiments were conducted using the xPC target machine running MATLAB 2020a Simulink Real-Time at a 500 Hz loop frequency. The thruster PWM-to-force mapping was characterized before the experiments and is summarized in Table~\ref{tab:thrust-table}. The SLIP parameters are described as follows $l_0 = 0.55~\text{m}, k_s = 1380~\text{N}/\text{m},$ and $m_b = 7.1~\text{kg}$. The spring stiffness was selected such that a nominal compression of 0.05 m produces a GRF approximately equal to body weight, ensuring the SLIP model operates within a physically meaningful force limit during stance. The QP-MPC runs at a rate of 500 Hz with $\text{N} = 5$ horizon. The friction coefficient was set as $\mu = 0.5$. The lower and upper bounds for decision variables are $[-8.7,10,-75,0]^{\top}$ and $[8.7,10,75,150]^{\top}$ respectively. The cost matrices for states and inputs are defined as $Q = diag([100,40,5000,1000])$ and $R = diag([0.05,0.3,1,1])$. Position tracking cost is intentional relaxed to accommodate the limit forward thrust authority.

\subsection{Experimental Results \& Discussion}
Figure~\ref{fig:-hardware-trot} presents the stable trotting results snapshot from Experiment B. The body position, velocity, and orientation are calculated using the onboard EKF estimator that combined IMU data at 500 Hz and OptiTrack data at 240 Hz. The thrust and stance leg was controlled using the proposed controller, while the swing leg was driven to the desired location using Raibert's heuristic foot placement controller. 

Figure~\ref{fig:-qp_solver_info} illustrates the computation performance of the QP solver on hardware. The total computation time for both experiments, which includes the setup-time and solve-time, consistently remain below $0.2\text{ms}$, well within the $2\text{ms}$ control loop. Additionally, the solver converges in max 10 iteration, confirming the QP formulation remains well-conditioned throughout the locomotion. 

Figure~\ref{fig:-grf} shows the QP-commanded GRF distribution within friction cone bounds of $\mu = 0.5$ for both experiments. The QP controller was successfully able to modulate the thrust force while enforcing the no-slip contact constraint across both experiments. The time-colored trajectory confirms a consistent GRF regulation with no degradation over the course of the experiments.

Figure~\ref{fig:-thrust} presents the thruster force commanded during both experiments. The QP-commanded forward thrust frequently approaches the hardware saturation limit of $\pm 8.7$N, indicating that the optimizer consistently requests forces near the physical capability of the thruster system. This is attributed to Thrust-to-weight contribution of the forward thruster, which produce a max of $8.7\text{N}$ compared to the robot's $70~\text{N}$ body weight. Despite operating near saturation for extended periods, the QP-TSLIP controller maintains stable locomotion throughout the experiments as evidenced by the stable orientation response.

The vertical thruster force is included as an optimization variable within the QP but is bounded to a constant value of $10\text{N}$ which is equally distributed to both the side thrusters. This design choice is motivated by safety considerations specific to thruster-assisted biped locomotion - aggressive variation in vertical thrust risk inducing unintended floating behavior, since high $u_{t,z}$ combined with $u_{t,r}$ could cause sudden roll disturbance on the swing side. By constraining $u_{t,z}$ to a fixed gravity compensation value within the optimizer, the QP retains a consistent formulation while ensuring predictable ground contact dynamics throughout locomotion. The left and right side thruster forces combine the QP vertical force baseline of $5\text{N}$ per thruster with PID orientation correction. The side thruster oscillates dynamically between $5- 25\text{N}$ as the orientation controller actively rejects roll disturbances throughout locomotion.

Figure~\ref{fig:-states} presents the COM body state trajectory for both experiments. The forward velocity reference is constant at $0~\text{m}/\text{s}$ and $0.1~\text{m}/\text{s}$ for Experiment A and Experiment B respectively. The instantaneous velocity oscillates due to thrust force tracking delay, and impact forces of the swing-to-stance transitions. The mean forward velocity remains close to its respective reference. The forward position reference is re-initialized at each touchdown to the current robot position and marched forward at the commanded velocity, preventing position error accumulation across steps. The pitch and roll angles remain bounded with $\pm 9^{\circ}$ and $\pm 4^{\circ}$ respectively, confirming stable sagittal and frontal-plane dynamics under the proposed decoupled control architecture. Fig.~\ref{fig:-traj-track-err} presents the joint tracking errors for all six joints. Joint tracking errors remain bounded within $\pm 1^{\circ}$ across both experiments.

Figure~\ref{fig:-mpc-prediction} validates the linearized SLIP model used within the QP framework. At each control cycle, the current robot state is used as an initial condition for a 5-step open-loop rollout using the QP-optimal inputs. The prediction rollout qualitatively captures the trend of the hardware trajectory. Since jerk is not included as a state variable in the ROM, sudden velocity changes arising from contact transition and impact dynamics which are not captured within the prediction model. As a result, brief deviations between the prediction and actual trajectories are expected at the swing-to-stance phase transition, where impulsive GRF produces rapid velocity changes that lie outside the modeling assumptions of the linearized SLIP dynamics. The frequent re-initialization inherent to the receding horizon strategy effectively prevents error accumulation.
\section{CONCLUSIONS}
This paper presented the QP-TSLIP controller for thruster-assisted bipedal locomotion on the hardware platform. The proposed decoupled architecture handles sagittal locomotion via a condensed QP-MPC with SLIP constraint, while frontal plane stability is maintained by a cascade controller. Hardware experiments across 25 steps demonstrate real-time QP solving with 0.18 ms at 500 Hz, consistent GRF cone satisfaction, and stable pitch and roll within $\pm 9^{\circ}$ and $\pm 4^{\circ}$ respectively. 

Future work will focus on three directions: extending the formulation to an actuated SLIP model enabling active spring length modulation for more aggressive locomotion; incorporating explicit thruster tracking delay into the QP constraint to further reduce the velocity tracking error, and relaxing the constant vertical thrust bound to enable adaptive z thrust for more aggressive locomotion, such as $1-2$m high jumps.










\printbibliography

\end{document}